\documentclass{article}

\usepackage[final]{neurips_2025}

\usepackage[utf8]{inputenc} 
\usepackage[T1]{fontenc}    
\usepackage{hyperref}       
\usepackage{url}            
\usepackage{booktabs}       
\usepackage{amsfonts}       
\usepackage{nicefrac}       
\usepackage{microtype}      
\usepackage{xcolor}         
\usepackage[table,dvipsnames]{xcolor}
\usepackage{booktabs}       
\usepackage{multirow}       

\usepackage{amsmath}
\usepackage{amssymb}
\usepackage{mathtools}
\usepackage{amsthm}
\usepackage{relsize}
\usepackage[capitalize,noabbrev]{cleveref}
\usepackage{wrapfig}
\usepackage{soul} 
\usepackage{paralist}  
\usepackage[most]{tcolorbox}
\usepackage{algorithm}
\usepackage[noend]{algpseudocode}

\DeclareMathSymbol{\dv}{\mathord}{operators}{"3A}

\newcommand*\raiseup[1]{%
        \begingroup
        \setbox0\hbox{\scriptsize\strut #1}%
        \dimen0=\ht\strutbox
        \advance\dimen0 by -\ht0
        \divide\dimen0 by 2
        \leavevmode
        \raise\dimen0\box0
        \endgroup
}

\usepackage[dvipsnames]{xcolor}      
\usepackage{soul}        
\usepackage{xcolor}      

\title{RELATE: A Schema-Agnostic Perceiver Encoder for Multimodal Relational Graphs}

\author{%
Joe Meyer$^{*}$ \\
SAP \\
Palo Alto, CA, USA \\
\texttt{joseph.meyer@sap.com}
\And
Divyansha Lachi$^{*,\dagger}$ \\
University of Pennsylvania \\
Philadelphia, PA, USA \\
\texttt{div11@upenn.edu}
\AND
Mahmoud Mohammadi \\
SAP \\
Seattle, WA, USA \\
\texttt{reza.mohammadi@sap.com}
\And
Roshan Reddy Upendra \\
SAP \\
Palo Alto, CA, USA \\
\texttt{roshan.reddy.upendra@sap.com}
\And
Eva L. Dyer \\
University of Pennsylvania \\
Philadelphia, PA, USA \\
\texttt{eva.dyer@upenn.edu}
\And
Mark Li \\
SAP \\
Seattle, WA, USA \\
\texttt{mark.li01@sap.com}
\And
Tom Palczewski \\
SAP \\
Palo Alto, CA, USA \\
\texttt{tom.palczewski@sap.com}
}

\begin{document}

\begingroup
\renewcommand\thefootnote{}
\footnotetext{$^*$Equal contribution, corresponding authors}
\footnotetext{$^\dagger$Work done during internship.}
\endgroup

\maketitle

\begin{abstract}
Relational multi-table data is common in domains such as e-commerce, healthcare, and scientific research, and can be naturally represented as heterogeneous temporal graphs with multi-modal node attributes. Existing graph neural networks (GNNs) rely on schema-specific feature encoders, requiring separate modules for each node type and feature column, which hinders scalability and parameter sharing. We introduce \textbf{RELATE} (Relational Encoder for Latent Aggregation of Typed Entities), a schema-agnostic, plug-and-play feature encoder that can be used with any general purpose GNN. RELATE employs shared modality-specific encoders for categorical, numerical, textual, and temporal attributes, followed by a Perceiver-style cross-attention module that aggregates features into a fixed-size, permutation-invariant node representation. We evaluate RELATE on ReLGNN and HGT in the RelBench benchmark, where it achieves performance within 3\% of schema-specific encoders while reducing parameter counts by up to 5x. This design supports varying schemas and enables multi-dataset pretraining for general-purpose GNNs, paving the way toward foundation models for relational graph data.
\end{abstract}

\section{Introduction}
\label{sec:intro}

Learning from relational multi-table data is a core challenge in domains such as e-commerce, healthcare, finance, and scientific discovery~\cite{codd1970relational}. This data can be naturally represented as heterogeneous temporal graphs, where nodes and edges have different types and attributes span multiple modalities—including text, time-series, and numerical values. Effectively modeling such graphs requires handling diverse schemas and capturing high-dimensional, multimodal inputs associated with each node type.

Recent graph neural networks (GNNs) such as HGT~\cite{hu2020heterogeneous} and RelGNN~\cite{robinson2024relbench} have shown promising results on individual relational datasets. However, these models are tightly coupled to the underlying schema: they require separate encoders for each node type and feature column, leading to architectures that (i) scale poorly with the number of columns, (ii) incur high memory and parameter costs, and (iii) inhibit generalization to new datasets with unseen schemas. As a result, these models are not well-suited for foundation model training, where a single model must handle diverse and non-aligned feature spaces across datasets.

To address these challenges, we introduce the \textbf{RELATE} (Relational Encoder for Latent Aggregation of Typed Entities), a schema-agnostic feature encoder designed for heterogeneous graphs with multimodal node features. RELATE uses modality-specific modules shared across all columns of the same type (e.g., categorical, numerical, textual), followed by a Perceiver-style cross-attention layer \cite{jaegle2021perceiver, graphfm} that compresses the set of column embeddings into a compact, fixed-size representation per node. This design is permutation-invariant to column order, accommodates varying schemas across node types, and scales to datasets with hundreds of features by attending from a small number of learnable latent queries.

We evaluate RELATE on a wide range of node classification and a subset of regression tasks from the RelBench benchmark~\cite{robinson2024relbench}, comparing against schema-specific encoders used in prior RDL frameworks.

RELATE achieves performance on par (within 3\%) with task-specific encoders on while reducing parameter count in datasets with a large number of features. Its schema-agnostic design enables plug-and-play integration into existing GNN architectures and paves the way toward scalable multi-dataset pretraining for general-purpose graph foundation models.

\vspace{0.5em}
\textbf{Our contributions are as follows:}
\begin{itemize}
\vspace{-0.5em}
\item We propose \textbf{RELATE}, a schema-agnostic encoder for heterogeneous temporal graphs that replaces per-column and per-type encoders with shared modality-specific modules, and uses a Perceiver-style cross-attention bottleneck to summarize variable-length column embeddings into fixed-size node representations.
\item RELATE generates fixed-size node embeddings that integrate seamlessly with existing GNN architectures (e.g., HGT, RelGNN), enabling plug-and-play use.
\item We show that RELATE achieves strong performance across diverse tasks in the RelBench benchmark—within 3\% of task-specific encoders for classification—while reducing parameter count by up to \textit{5x}.
\end{itemize}

\begin{figure}[t!]
    \centering
    \includegraphics[width=1.01\textwidth]{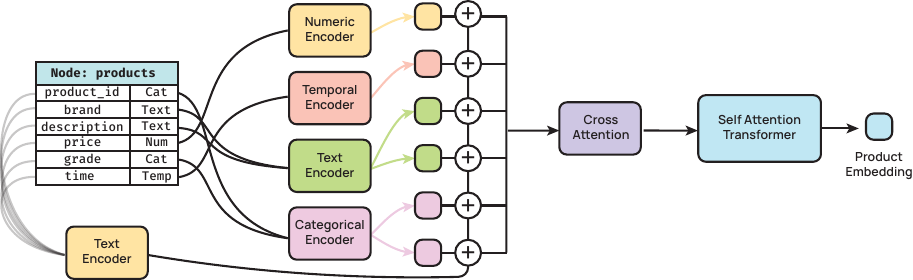}
    \caption{\footnotesize{\textbf{Overview of the RELATE architecture.} Columns are encoded by shared modality-specific encoders, aggregated with column metadata, and summarized by a cross-attention transformer into fixed-size node embeddings for GNNs. Please refer to Appendix~\ref{app:col_agg} for details on how column metadata is aggregated with cell embeddings.}} \vspace{-3mm} 
\label{fig:overview}
    
\end{figure}

\section{Background}
\label{sec:background}

Relational databases are widely used in domains such as finance, healthcare, and human resources. Traditional pipelines rely on costly joins to flatten relational data into a single feature matrix. Relational Deep Learning (RDL) instead exploits the relational schema directly, learning from interconnected tables without materializing large joins~\cite{robinson2024relbench}.  

\paragraph{From relational data to graphs.}
Let a database be $(\mathcal{T}, \mathcal{R})$, where $\mathcal{T}=\{T_1,\dots,T_n\}$ is a set of tables and $\mathcal{R}\subseteq \mathcal{T}\times\mathcal{T}$ encodes foreign-key $\rightarrow$ primary-key relations. Each table contains entities (rows) with identifiers, foreign keys, features, and timestamps. This structure is naturally modeled as a heterogeneous temporal graph
$\mathcal{G}=(\mathcal{V},\mathcal{E},\phi,\psi,\tau)$, where nodes $\mathcal{V}$ are entities, directed edges $\mathcal{E}\subseteq \mathcal{V}\times\mathcal{V}$ represent relations, $\phi:\mathcal{V}\!\to\!\mathcal{T}_V$ and $\psi:\mathcal{E}\!\to\!\mathcal{T}_E$ assign node/edge types, and $\tau$ attaches timestamps to nodes or edges.

\paragraph{The standard encoder in RDL.}
The common practice is to give \emph{each feature column} its own encoder and then concatenate the resulting embeddings. For a node $v$, let $\mathcal{C}_v$ be its set of feature columns. Each column $c\in\mathcal{C}_v$ is mapped to an embedding $\phi_c(\mathbf{x}_v^{(c)})$, where $\mathbf{x}_v^{(c)}$ is the raw value(s) of column $c$ and $\phi_c$ is a \emph{column-specific} encoder selected by modality (e.g., MLP for numeric, embedding lookup for categorical). The initial node representation is
\[
\mathbf{h}_v^{\text{concat}} \;=\; \bigoplus_{c\in\mathcal{C}_v}\, \phi_c\!\bigl(\mathbf{x}_v^{(c)}\bigr),
\]
with $\oplus$ denoting concatenation. A tabular backbone $f_{\text{tab}}$ (often a ResNet~\cite{hu2024pytorch}) is then applied before passing the result to a downstream GNN.

    

\paragraph{Limitations.}
While expressive, this design tightly couples the model to each schema: every new column introduces a new encoder ($\phi_c$), and every node type typically requires its own backbone. This leads to parameter explosion in databases with hundreds of columns and hinders schema-agnostic pretraining across datasets. Related homogeneous GFMs (e.g., GraphFM~\cite{graphfm}) also rely on dataset-specific MLPs~\cite{rosenblatt1957perceptron}, limiting transfer to new schemas.

\section{Method}
\label{sec:method}

We propose \textbf{RELATE}, a schema-agnostic encoder for heterogeneous temporal graphs with high-dimensional, multimodal node features. RELATE is a plug-and-play feature encoder that enables multi-dataset pretraining for any general-purpose GNN model. RELATE consists of two key components: (i) a library of modality-specific encoders shared across all node types and columns, and (ii) a Perceiver-style cross-attention module that aggregates a variable-length set of column embeddings into a fixed-size embedding  which serves as the initial node representation for the model.

\subsection{Modality-Specific Encoders}
\label{sec:modality-encoders}

We group all node features into four high-level modalities and apply a shared encoder for each: (i) \textbf{Numerical}, representing continuous scalar values such as age or price; (ii) \textbf{Timestamp}, capturing time-stamped information such as event times or birth dates; (iii) \textbf{Categorical}, covering discrete non-text attributes such as product category or gender; and (iv) \textbf{Textual}, handled by a pretrained text encoder for free-text fields or high-cardinality categorical attributes expressed in text. Features from each modality are processed by a dedicated encoder shared across all columns of that type. These encoders are conditioned on the column metadata through a shared text embedding model.

\paragraph{Numerical Encoder.} We use the Fourier Number Embedding (FoNE) encoder~\cite{zhou2025fone} to map continuous scalar values into dense embeddings. Missing values (e.g., \texttt{NaN}) are represented by a learnable token, and unlike standard approaches, FoNE does not require normalization. An optional shared linear projection maps the resulting embedding into the target space. By design, FoNE is schema-independent and can be applied to any numeric feature.

\paragraph{Timestamp Encoder.}
Each timestamp is decomposed into interpretable components (year, month, day, etc.), then encoded using a combination of positional encodings (for absolute time) and cyclic encodings (for periodicity). A shared linear projection maps the encoded time features to the target embedding space.

\paragraph{Categorical Encoder.}
Categorical inputs are hashed into a shared vocabulary space, with the hash function conditioned on the column embedding. This allows identical values from different columns (e.g., “1”) to map to distinct embeddings, ensuring both efficient use of the embedding space and semantic separation across columns. The categorical encoder is suited for anonymized fields such as hashed IDs common in real-world databases, while features with semantically meaningful values are instead processed by a pretrained text encoder.

\paragraph{Pretrained Text Encoder.}
Textual columns are encoded with a pretrained sentence encoder~\cite{minishlab2024model2vec}, and a shared linear projection maps the resulting embeddings into the target dimension.

\subsection{Column-Level Metadata Conditioning}
To improve generalization across columns and datasets, we incorporate column-level metadata into each modality encoder. Column names, table names (node types), and optional descriptions are encoded with a pretrained text embedding model and injected into the feature encoding process, enabling the encoder to distinguish between semantically different columns that share the same modality and value space and to generalize to unseen schemas. We adopt different strategies for aggregating column metadata with cell embeddings depending on the modality; details are provided in Appendix~\ref{app:col_agg}.

\subsection{Permutation-Invariant Column Aggregation}
\label{sec:perceiver-aggregation}

To transform a variable-length set of column embeddings into a fixed-size node representation, we adopt a cross-attention module inspired by PerceiverIO~\cite{jaegle2021perceiver, graphfm}. A shared sequence of $L$ learnable latent tokens 
$\mathbf{Z} = [\mathbf{z}_1, \ldots, \mathbf{z}_L] \in \mathbb{R}^{L \times d}$ 
serves as queries, while the column embeddings for node $v$, 
$\mathbf{X}_v = [\mathbf{x}_1, \ldots, \mathbf{x}_{C_v}] \in \mathbb{R}^{C_v \times d}$, 
form the input sequence. Since $L \ll C_v$, this formulation reduces computational cost by decoupling self-attention from the number of input columns.  

\begin{equation}
    \label{eq:cross_attn}
    \mathbf{Z}_v \;=\;
    \mathbf{Z} +
    \mathrm{softmax}\!\left(\frac{\mathbf{Q}\mathbf{K}_v^\top}{\sqrt{d_k}}\right)\mathbf{V}_v,
\end{equation}

where $\mathbf{Q} = \mathbf{W}_q \mathbf{Z}$, 
$\mathbf{K}_v = \mathbf{W}_k \mathbf{X}_v$, 
and $\mathbf{V}_v = \mathbf{W}_v \mathbf{X}_v$. 
This operation is permutation-invariant to the input column order and enables flexible adaptation to nodes with different schemas and numbers of attributes. 
Following this compression, $N$ layers of self-attention are applied to the latent tokens to produce the final node representation $\mathbf{z}_v$.

\noindent \textbf{Remark.} 
Permutation invariance is especially valuable when training across datasets with overlapping but non-identical schemas. The same semantic node type (e.g., \texttt{user}, \texttt{item}) may appear with columns in different orders or with partial feature overlap. By avoiding reliance on fixed input positions, our aggregation mechanism produces consistent node representations across schemas, enabling schema-agnostic pretraining.
\section{Results}
\label{sec:results}

We evaluate RELATE on RelBench~\cite{robinson2024relbench}, a recently introduced benchmark for relational deep learning that spans seven real-world multi-table datasets across domains such as e-commerce, healthcare, and finance. These datasets are structured as heterogeneous temporal graphs with multiple node and edge types and rich, multimodal attributes. Our experiments cover two primary tasks: node classification and node regression. We report Area Under the ROC Curve (AUC) for classification and Mean Absolute Error (MAE) for regression. To ensure a fair comparison, all models are trained using the same splits and optimization protocols defined in the benchmark~\cite{robinson2024relbench}. We integrate RELATE into two widely-used architectures—Heterogeneous Graph Transformer (HGT)~\cite{hu2020heterogeneous} and RelGNN~\cite{robinson2024relbench}—demonstrating compatibility with standard backbones (refer to Appendix~\ref{app:baseline_models} for more details). We compare against the default heteroencoder used in RelBench~\cite{hu2024pytorch}, which uses distinct encoders for each node type and column. RELATE instantiates only a single backbone encoder across all node types and shares weights for each modality encoder across node types.

\subsection{Experimental Setup}

We implement RELATE within the RDL pipeline~\cite{robinson2024relbench} by replacing the original heteroencoder with our architecture. We preserve the underlying task logic and training infrastructure. RELATE is trained using the AdamW optimizer~\cite{loshchilov2017decoupled} with a fixed learning rate of $5 \times 10^{-3}$. 
All other settings, such as batch size and dropout remain fixed across datasets. Similarly, we fix the embedding dimension to 128.  Rather than perform exhaustive hyperparameter tuning, we examine the architecture's ability to learn in variable schemas. More information can be found in the appendix in \ref{sec:model_details}

\subsection{Performance Comparison}
\vspace{-3mm}

\begin{table*}[t]
\centering
\caption{\textbf{Results on RelBench for Standard Encoder v/s RELATE} We evaluate HGT and RelGNN with two feature encoders: Standard and RELATE. Classification uses AUC (higher is better); regression uses MAE (lower is better). $\Delta$ is RELATE$-$Standard.}
\label{tab:relbench}
\scriptsize
\setlength{\tabcolsep}{4pt}
\renewcommand{\arraystretch}{1.1}
\begin{tabular}{l l
                c c c
                c c c}
\toprule
\multirow{2}{*}{Dataset} & \multirow{2}{*}{Task} &
\multicolumn{3}{c}{\textbf{HGT}} & \multicolumn{3}{c}{\textbf{RelGNN}} \\
\cmidrule(lr){3-5} \cmidrule(lr){6-8}
 &  & Standard & RELATE & $\Delta$ (RELATE$-$Standard) & Standard & RELATE & $\Delta$ (RELATE$-$Standard) \\
\midrule
\multicolumn{8}{l}{\textbf{Classification — AUC }} \\
\midrule
rel-f1     & driver-dnf          & 0.7337 & 0.6653 & -0.0684 & 0.7135 & 0.6892 & -0.0243 \\
rel-f1     & driver-top3         & 0.8297 & 0.4779 & -0.3518 & 0.7701 & 0.6901 & -0.0800 \\
rel-avito  & user-clicks         & 0.6490 & 0.6434 & -0.0056 & 0.6590 & 0.6610 &  0.0020 \\
rel-avito  & user-visits         & 0.6459 & 0.6262 & -0.0197 & 0.6620 & 0.6625 &  0.0005 \\
rel-event  & user-repeat         & 0.7351 & 0.7234 & -0.0117 & 0.7551 & 0.6710 & -0.0841 \\
rel-event  & user-ignore         & 0.8130 & 0.8510 &  0.0380 & 0.8035 & 0.8114 &  0.0079 \\
rel-trial  & study-outcome       & 0.6695 & 0.5979 & -0.0716 & 0.6742 & 0.5838 & -0.0904 \\
rel-amazon & user-churn          & 0.6384 & 0.6548 &  0.0164 & 0.7033 & 0.6886 & -0.0147 \\
rel-amazon & item-churn          & 0.7529 & 0.7515 & -0.0014 & 0.8283 & 0.8124 & -0.0159 \\
rel-stack  & user-engagement     & 0.8723 & 0.8805 &  0.0082 & 0.9059 & 0.9010 & -0.0049 \\
rel-stack  & user-badge          & 0.8227 & 0.8226 & -0.0001 & 0.8890 & 0.8664 & -0.0226 \\
rel-hm     & user-churn          & 0.6561 & 0.6525 & -0.0036 & 0.6955 & 0.6937 & -0.0018 \\
\midrule
\multicolumn{2}{l}{\emph{Average $\Delta$ (AUC)}} &  &  & \textbf{-0.0393} &  &  & \textbf{-0.0274} \\
\midrule
\multicolumn{8}{l}{\textbf{Regression — MAE }} \\
\midrule
rel-f1     & driver-position     & 4.6649 & 6.1843 &  1.5194 & 4.2056 & 4.2621 &  0.0565 \\
rel-avito  & ad-ctr              & 0.0365 & 0.0382 &  0.0017 & 0.0424 & 0.0421 & -0.0003 \\
rel-trial  & site-success        & 0.4244 & 0.4353 &  0.0109 & 0.3457 & 0.4191 &  0.0734 \\
rel-stack  & post-votes          & 0.0679 & 0.0679 &  0.0000 & 0.0652 & 0.0649 & -0.0003 \\
rel-hm     & item-sales          & 0.0677 & 0.0708 &  0.0031 & 0.0556 & 0.0589 &  0.0033 \\
\midrule
\multicolumn{2}{l}{\emph{Average $\Delta$ (MAE)}} &  &  & \textbf{0.3070} &  &  & \textbf{0.0265} \\
\bottomrule
\end{tabular}
\end{table*}

On average, RELATE achieves accuracy within 3\% of dataset-specific encoders for HGT and RelGNN respectively on classification tasks, despite using a shared, schema-agnostic architecture. Table~\ref{tab:relbench} reports per-task metrics (AUROC or MAE).

\begin{figure}[t!]
    \centering
    \includegraphics[width=\linewidth]{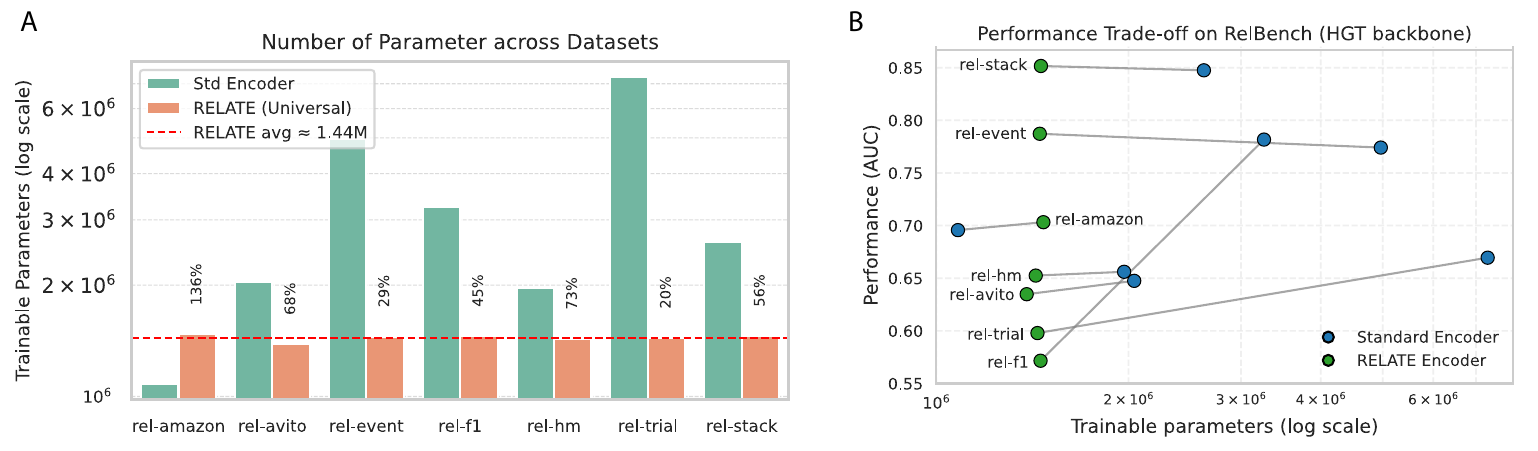}
    \caption{
\textbf{(A)} Parameter comparison across RelBench datasets. We compare the number of trainable parameters between schema-specific encoders (Std) and RELATE. Schema-specific encoders grow with the number of tables and features, whereas RELATE maintains a nearly constant footprint. Percentages above the RELATE bars indicate the parameter ratio relative to schema-specific encoders. 
\textbf{(B)} Performance v/s parameter count (log scale) for the HGT backbone. Each line connects a schema-specific encoder (blue) to its RELATE counterpart (green), illustrating the change in both performance and model size. RELATE achieves comparable AUC while using up to 5× fewer parameters.}
    \label{fig:param_eff}
    \vspace{-3mm}
\end{figure}

RELATE matches or slightly underperforms the RelBench heteroencoder on most tasks, while using a single backbone encoder across all node types and not being explicitly tied to any schema. Additionally, the number of learnable parameters is significantly smaller on several tasks, especially those with a large number of tables and features (Figure~\ref{fig:param_eff}B). For example, RELATE outperforms the comparison method on rel-event user-ignore while using only around 29\% of learnable parameters compared to the heteroencoder. We find that RELATE performs generally on par with dataset-specific encoders on regression, however in the rel-f1 task RELATE under performs by a larger margin. We expect this is due to rel-f1 having a small number of train examples and a trade-off of single dataset performance and universality.




\subsection{Parameter Efficiency}

Figure~\ref{fig:param_eff}A compares the number of trainable parameters between schema-specific encoders and RELATE across RelBench datasets (see \cref{tab:encoder_params_std_univ} in \cref{app:encoder_params} for exact parameter counts) . RELATE maintains a nearly constant parameter footprint across datasets, since it uses a fixed set of modality-specific encoders shared across node types. In contrast, the parameter count of schema-specific encoders grows rapidly with the number of tables and features. As a result, RELATE achieves up to a \textbf{5× reduction in parameters} on feature-rich datasets such as \textit{rel-trial} and \textit{rel-event}, while remaining competitive in cases with fewer features. This stability makes RELATE particularly advantageous for real-world schemas involving hundreds or thousands of attributes, where schema-specific encoders become prohibitively large. Figure~\ref{fig:param_eff}B further highlights the trade-off between performance and parameter efficiency. In datasets such as \textit{rel-hm} and \textit{rel-avito}, RELATE maintains similar performance while using significantly fewer parameters.

\begin{wrapfigure}{r}{0.45\textwidth}
    \centering
   \includegraphics[width=\linewidth]{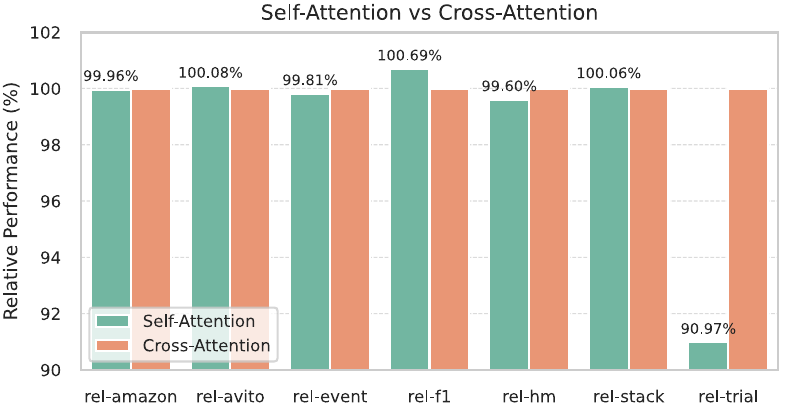}
   \vspace{-4mm}
\caption{\footnotesize{\textbf{Ablation}: Attention ablation for RelGNN, reporting average AUC performance relative to cross-attention. Bars show the performance of self-attention compared to cross-attention within each dataset.}}
    \label{fig:attn_abl}
    \vspace{-4mm}
\end{wrapfigure}


\subsection{Ablation Study}

We evaluate the role of cross-attention by replacing it with a full self-attention (SA) mechanism applied over the input tokens. 
As shown in Figure~\ref{fig:attn_abl}, we report the relative performance of self-attention for RelGNN, averaged over all classification tasks. 
This substitution yields marginal gains on a few datasets but does not consistently improve performance, with self-attention achieving only 90--101\% of cross-attention on average. 
At the same time, the computational overhead of full self-attention is substantially higher due to its quadratic complexity. 
In contrast, the cross-attention bottleneck in RELATE provides a more efficient and scalable alternative, preserving accuracy while enabling training on large and feature-rich graphs. 
Comprehensive AUC and MAE results for both RelGNN and HGT are reported in Appendix~\ref{app:attention_ablation}.

\section{Related Work}
\label{sec:related_work}

\paragraph{Tabular Foundation Models}Recent models such as ConTextTab~\citep{spinaci2025contexttab} and PORTAL~\citep{spinaci2024portal} leverage table semantics, e.g., column headers, as context for cell values. Ablations showed that removing headers causes significant drops in performance, highlighting the importance of semantic metadata. These models typically use shared, modality-specific encoders for multi-modal values (numerical, categorical, text, temporal). ConTextTab, for example, scales numeric values, applies a learnable vector, encodes text with a pretrained language model, and sums embeddings for temporal components. Other approaches such as XTAB~\citep{zhu2023xtab} pretrain only the transformer backbone while relying on dataset-specific feature preprocessing, which limits transfer across schemas.

\paragraph{Graph Foundation Models}One of the key challenges of GFMs are to design architectures that can transfer across varying input spaces \cite{mao2024position}. Recent homogeneous GFM models~\citep{graphfm} demonstrate rapid transfer to downstream tasks after pretraining, but still rely on dataset-specific featurization that limits generalization and prevents true zero-shot transfer. RELATE addresses these limitations by introducing task-agnostic encoders that integrate directly with standard GNN backbones. The concurrent work Griffin~\citep{wang2025griffin} uses pretrained encoders for text and numeric features and applies cross-attention over cells, column metadata, and task information. Its feature encoders are task-conditioned and limited to text and numeric modalities, potentially conflating categorical IDs with text. Other GFMs, such as OFA \cite{liu2023oneforall}, only operate on text-attributed graphs (TAGs). GraphAlign \cite{hou2024graphalign} aims to align feature distributions across across diverse graphs by leveraging mixture of experts (MOE), however the model is similarly limited to TAGs.

\section{Conclusion}
\label{sec:conclusion}

We introduced RELATE, a schema-agnostic encoder for heterogeneous temporal graphs that replaces per-column and per-type feature stacks with shared modality-specific modules and a Perceiver-style cross-attention bottleneck. By attending from a fixed set of latent tokens to variable-length column embeddings, RELATE achieves permutation invariance to column order and decouples the complexity of self-attention from the raw feature dimensionality of a node. We integrate RELATE with widely used heterogeneous GNN backbones, including HGT and RelGNN, demonstrating its effectiveness as a plug-and-play encoder that achieves competitive accuracy while substantially reducing parameter counts. Beyond reducing redundancy across schemas, RELATE provides a foundation for learning across datasets and may serve as a building block for future multi-dataset pretraining on general-purpose GNNs. Future work aims to evaluate RELATE in such multi-dataset settings, analyze how schema diversity and structural variation influence generalization, and conduct systematic ablations to disentangle the contributions of architectural components and training design choices.


\section{Acknowledgments and Disclosure of Funding}
We thank Alexandre Dorais, Jialin Dong, Viswa Ganapathy, Dinesh Katupputhur, Prasanna Lalingkar, Karan Paresh, Andrew Pouret, Afreen Shaikh, and Jay Shinigari for their valuable assistance in conducting the experiments and Qiunan Ma for helping with the figures. This project was also supported by NSF CAREER Award RI:2146072, NSF
award CIF:RI:2212182 as well as generous gifts from the CIFAR Azrieli Global Scholars Program (D.L and E.L.D).

\newpage
{\small
\bibliographystyle{abbrv}
\bibliography{graphbib}
}

\newpage

\appendix
\onecolumn
\section{Appendix}

\subsection{Model Details}
\label{sec:model_details}
\subsubsection{Hyperparameters}

For both the full self-attention and cross-attention encoder variants, we fix the number of attention heads and layers to 4, apply a dropout rate of 0.2, and set the hidden dimension to 128. In the Perceiver encoder, the number of latent tokens is set to 8. All models are trained for up to 10 epochs. For the GNN backbone, we use 128 channels, a 2-hop neighborhood, and uniformly sample 128 neighbors per node.

\subsubsection{RELATE Column Meta Aggregation}
\label{app:col_agg}

We opt to aggregate column information for numeric and text columns by adding the multi-modal cell embedding to the projection of the column metadata (e.g., column name). Then, we input this into a two layer MLP with RELU activation. Specifically: 

\begin{equation}
\begin{aligned}
X &= \text{CellEmbeddings} \; W_{\text{shared}} \\
H &= \text{ColProj}\!\left(\text{ColumnEmbeddings}\right) \\
Z &= X + H \\
\tilde{Z} &= Z + \text{MLP}(Z), \quad 
\text{MLP}(Z) = \text{ReLU}(Z W_1 + b_1) W_2 + b_2
\end{aligned}
\end{equation}

For hashed features we do not incorporate column metadata. Finally, for time columns we find gating to be effective. We apply the sigmoid function to the projection of the column embeddings. We multiply the result with the projection of the time embeddings.

\begin{equation}
Z \;=\; \big( \text{CellEmbeddings}(W_{\text{shared}})) \;\odot\; \sigma\!\left( \text{ColProj}(\text{ColumnEmbeddings}) \right)
\end{equation}

\subsubsection{RELATE Cross-Attention}

Here we describe the cross-attention performed by RELATE. \textit{X} is the result of the multi-modal encoders described in \ref{sec:modality-encoders}. \textit{X} contains an embedding for each cell according to its modality. \textit{Q}, \textit{K} ,and \textit{V} are all projected using shared weights and \textit{L} is initialized as a learnable parameter which represents the latent tokens. After cross-attention is performed, self-attention is executed on the latents. This process is performed for \textit{N} layers.

\begin{equation}
\begin{aligned}
Q &= L W_Q, \quad L \in \mathbb{R}^{B \times N_{\text{lat}} \times d} \\
K &= X W_K, \quad V = X W_V, \quad X \in \mathbb{R}^{B \times N_{\text{cells}} \times d} \\[6pt]
\text{CrossAttn}(L, X) &= \text{softmax}\!\left( \frac{Q K^\top}{\sqrt{d}} \right) V \\[6pt]
L' &= L + \text{CrossAttn}(L, X) \\[6pt]
L'' &= L' + \text{SelfAttn}(L') \\[6pt]
\end{aligned}
\end{equation}

\subsection{Backbone Models}
\label{app:baseline_models}

\paragraph{RelGNN: }
The Relational Graph Neural Network (RelGNN)~\cite{robinson2024relbench} serves as a strong heterogeneous GNN baseline that extends message passing to relational databases represented as multi-entity graphs. Unlike traditional GNNs that operate on homogeneous graph structures, RelGNN performs message passing over relational graphs where nodes represent entities (e.g., tables or records) and edges encode typed relationships derived from foreign key links. This schema-aware formulation enables the model to jointly reason over multiple interconnected tables while preserving the structural and semantic integrity of the underlying database. In our experiments, we adopt the default hyperparameter configuration provided in RelBench~\cite{robinson2024relbench}.

\paragraph{HGT: }

The Heterogeneous Graph Transformer (HGT)~\cite{hu2020heterogeneous} extends the transformer architecture to handle multi-relational graphs with multiple node and edge types. Unlike standard graph transformers that assume homogeneous structures, HGT introduces type-specific transformation matrices and relation-aware attention to model heterogeneity in both message passing and aggregation. Each edge type is associated with a unique attention function, allowing the model to capture asymmetric relationships between different entity types.

In our experiments, we use the HGT+PE variant, which augments the standard HGT with Laplacian positional encodings (LapPE)~\cite{arora2025exploiting} computed over sampled subgraphs to incorporate structural context. We employ the HGTConv implementation from PyTorch Geometric~\cite{fey2024graph}, integrated into the RDL pipeline~\cite{robinson2024relbench}. The model consists of 2 HGT layers with 4 attention heads each, residual connections, and layer normalization. The dimensionality of the LapPE is set to 4 for most datasets and 2 for datasets such as \texttt{rel-amazon} and \texttt{rel-hm}. 

\subsection{Ablation Results}

\label{app:attention_ablation}

Table~\ref{tab:attention_ablation} reports the complete results of our attention ablation study, comparing the Perceiver-style cross-attention encoder with a full self-attention (SA) variant for both HGT and RelGNN across all RelBench datasets. 

For classification tasks, we evaluate performance using AUC (higher is better). On average, cross-attention outperforms self-attention by $+0.0068$ AUC in HGT and $+0.0039$ AUC in RelGNN. While self-attention achieves marginal improvements on a few tasks, these gains are inconsistent and come with a substantial increase in computational cost. 

For regression tasks, we report MAE (lower is better). Here, cross-attention again performs slightly better, reducing MAE by $0.0670$ in HGT and $0.0160$ in RelGNN on average. 

Overall, the results confirm that the latent cross-attention bottleneck provides accuracy on par with or better than full self-attention while offering substantially higher efficiency, making it the preferable choice for training on large and feature-rich graphs.

\begin{table*}[hbtp]
\centering
\caption{\textbf{Attention Ablation.} We compare full self-attention and cross-attention (Perceiver). Classification uses AUC (higher is better); regression uses MAE (lower is better). $\Delta$ is Perceiver$-$Full.}
\label{tab:attention_ablation}
\scriptsize
\setlength{\tabcolsep}{5pt}
\renewcommand{\arraystretch}{1.1}
\begin{tabular}{l l
                c c c
                c c c}
\toprule
\multirow{2}{*}{Dataset} & \multirow{2}{*}{Task} &
\multicolumn{3}{c}{\textbf{HGT}} & \multicolumn{3}{c}{\textbf{RelGNN}} \\
\cmidrule(lr){3-5} \cmidrule(lr){6-8}
 &  & Full & Perceiver & $\Delta$ (Perceiver$-$Full) & Full & Perceiver & $\Delta$ (Perceiver$-$Full) \\
\midrule
\multicolumn{8}{l}{\textbf{Classification — AUC }} \\
\midrule
rel-f1     & driver-dnf          & 0.6359 & 0.6653 &  0.0294 & 0.6661& 0.6892 &  0.0231 \\
rel-f1     & driver-top3         & 0.5609 & 0.4779 & -0.0830 & 0.7227& 0.6901 & -0.0326 \\
rel-avito  & user-clicks         & 0.6316 & 0.6434 &  0.0118 & 0.6618& 0.6610 & -0.0008\\
rel-avito  & user-visits         & 0.6272 & 0.6262 & -0.0010 & 0.6627& 0.6625 & -0.0002 \\
rel-event  & user-repeat         & 0.6036 & 0.7234 &  0.1198 & 0.6554& 0.6710 & 0.0156\\
rel-event  & user-ignore         & 0.8412 & 0.8510 &  0.0098 & 0.8242& 0.8114 & -0.0128\\
rel-trial  & study-outcome       & 0.5275 & 0.5979 &  0.0704 & 0.5311   & 0.5838 &  0.0527 \\
rel-amazon & user-churn          & 0.6727 & 0.6548 & -0.0179 & 0.6894   & 0.6886 & -0.0008 \\
rel-amazon & item-churn          & 0.7878 & 0.7515 & -0.0363 & 0.8110   & 0.8124 &  0.0014 \\
rel-stack  & user-engagement     & 0.8780 & 0.8805 &  0.0025 & 0.9016   & 0.9010 & -0.0006 \\
rel-stack  & user-badge          & 0.8236 & 0.8226 & -0.0010 & 0.8668   & 0.8664 & -0.0004 \\
rel-hm     & user-churn          & 0.6760 & 0.6525 & -0.0235 & 0.6909   & 0.6937 &  0.0028 \\
\midrule
\multicolumn{2}{l}{\emph{Average $\Delta$ (AUC)}} & & & \textbf{0.0068} & & & \textbf{0.0039}\\
\midrule
\multicolumn{8}{l}{\textbf{Regression — MAE }} \\
\midrule
rel-f1     & driver-position     & 6.5346 & 6.1843 & -0.3503 & 4.3267 & 4.2621 & -0.0645 \\
rel-avito  & ad-ctr              & 0.0387 & 0.0382 & -0.0005 & 0.0415& 0.0421 & 0.0006\\
rel-trial  & site-success        & 0.4241 & 0.4353 &  0.0112 & 0.4339   & 0.4191 & -0.0148 \\
rel-stack  & post-votes          & 0.0679 & 0.0679 &  0.0000 & 0.0651   & 0.0649 & -0.0002 \\
rel-hm     & item-sales          & 0.0664 & 0.0708 &  0.0044 & 0.0600   & 0.0589 & -0.0011 \\
\midrule
\multicolumn{2}{l}{\emph{Average $\Delta$ (MAE)}} & & & \textbf{-0.0670} & & & \textbf{-0.0160}\\
\bottomrule
\end{tabular}
\end{table*}


\subsection{Parameter Comparison}

\label{app:encoder_params}

Table~\ref{tab:encoder_params_std_univ} reports the exact parameter counts corresponding to Figure~\ref{fig:param_eff}. As shown, the parameter footprint of schema-specific encoders increases substantially with the number of tables and features in each dataset, ranging from $1.1\text{M}$ on \textit{rel-amazon} to over $7.3\text{M}$ on \textit{rel-trial}. In contrast, RELATE remains nearly constant at $\sim$1.4M parameters across all datasets, since it reuses a fixed set of modality-specific encoders. This property results in significant parameter savings—up to a \textbf{5× reduction} on feature-rich datasets such as \textit{rel-trial} and \textit{rel-event}—while still maintaining competitive performance.

\begin{table*}[hbtp]
\centering
\scriptsize
\setlength{\tabcolsep}{6pt}
\renewcommand{\arraystretch}{1.1}
\caption{\textbf{Parameter comparison:} We compare the number of trainable parameters between the heteroencoder and RELATE Standard across datasets.}
\label{tab:encoder_params_std_univ}
\begin{tabular}{llcccc}
\toprule
Dataset  &\# of Tables& \# of Features& Std. Encoder (\#params) & Universal (\#params) & Universal / Std (\%) \\
\midrule
rel-amazon  &3& 15& $1.08122 \times 10^{6}$ & $1.4713 \times 10^{6}$ & 136.078 \\
rel-avito   &8& 43& $2.04058 \times 10^{6}$ & $1.38483 \times 10^{6}$ & 67.8648 \\
rel-event   &5& 128& $4.96602 \times 10^{6}$ & $1.45178 \times 10^{6}$ & 29.2342 \\
rel-f1      &9& 77& $3.25939 \times 10^{6}$ & $1.45626 \times 10^{6}$ & 44.6788 \\
rel-hm      &3& 37& $1.968 \times 10^{6}$   & $1.4313 \times 10^{6}$  & 72.7285 \\
rel-trial   &15& 140& $7.29984 \times 10^{6}$ & $1.44026 \times 10^{6}$ & 19.73 \\
 rel-stack &7& 51& 2.62387e+06& $1.457672 \times 10^{6}$&55.50\\
 \bottomrule
\end{tabular}
\end{table*}

\end{document}